\documentclass[letterpaper,times, 10pt,twocolumn]{article}

\usepackage{latex12}
\usepackage{times}
\usepackage{balance}
\usepackage{subfigure}
\usepackage{graphicx}

\begin{document}

\title{Binary Stereo Matching}

\author{Kang Zhang, Jiyang Li, Yijing Li, Weidong Hu, Lifeng Sun, Shiqiang Yang\\
\emph{Department of Computer Science, Tsinghua University, Beijing, China}\\
}

\maketitle
\thispagestyle{empty}

\begin{abstract}
In this paper, we propose a novel binary-based cost computation and aggregation approach for stereo matching problem. The cost volume is constructed through bitwise operations on a series of binary strings. Then this approach is combined with traditional winner-take-all strategy, resulting in a new local stereo matching algorithm called binary stereo matching (BSM).  Since core algorithm of BSM is based on binary and integer computations, it has a higher computational efficiency than previous methods. Experimental results on Middlebury benchmark show that BSM has comparable performance with state-of-the-art local stereo methods in terms of both quality and speed. Furthermore, experiments on images with radiometric differences demonstrate that BSM is more robust than previous methods under these changes, which is common under real illumination.
\end{abstract}

\vspace{-20pt}
\Section{Introduction}
\label{sec:intro}
\vspace{-8pt}

\begin{figure}[t]
  \centering
  \subfigure[]{\label{fig:cones_l}\includegraphics[width=0.1\textwidth]{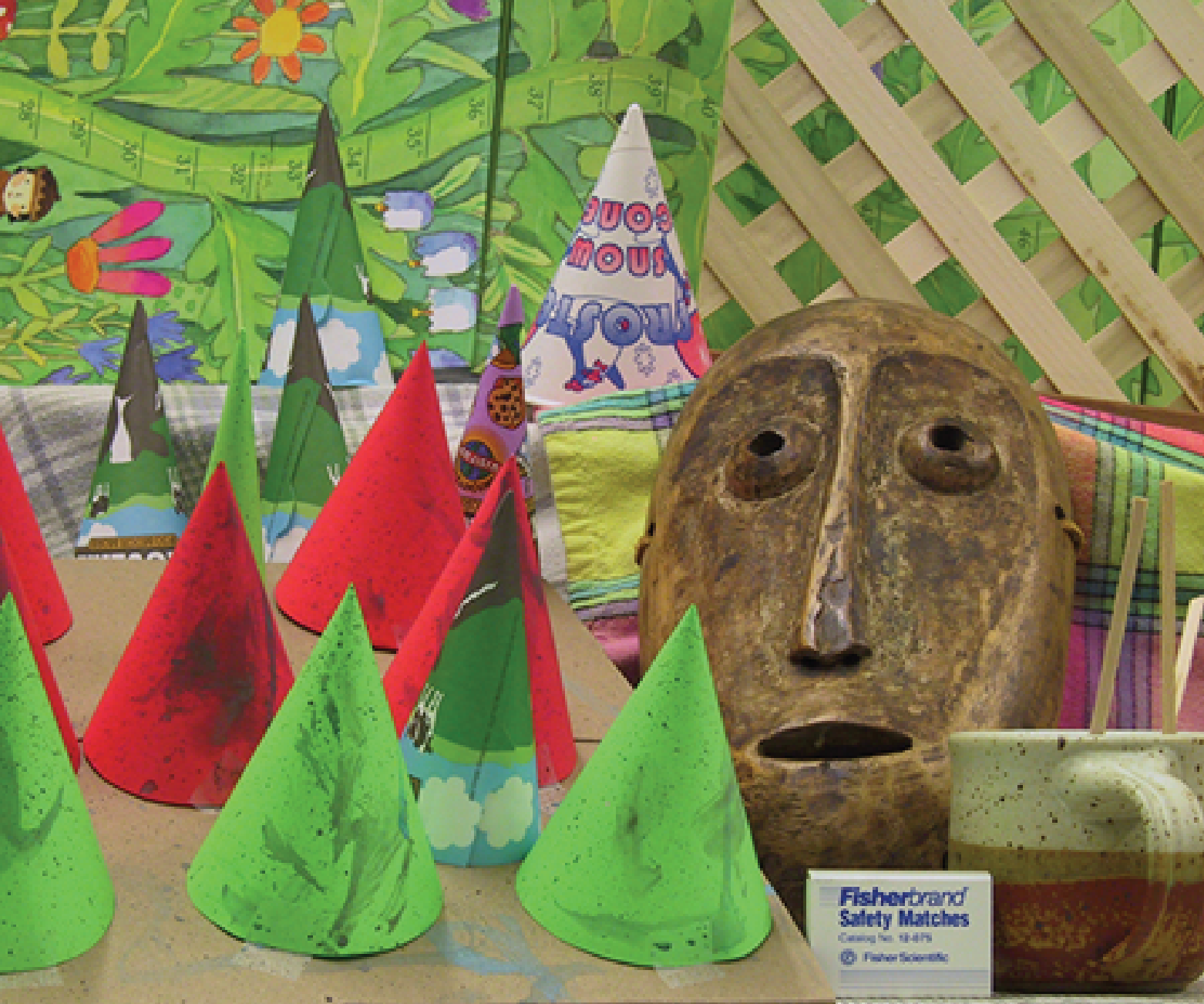}}
  \subfigure[]{\label{fig:cones_noca}\includegraphics[width=0.1\textwidth]{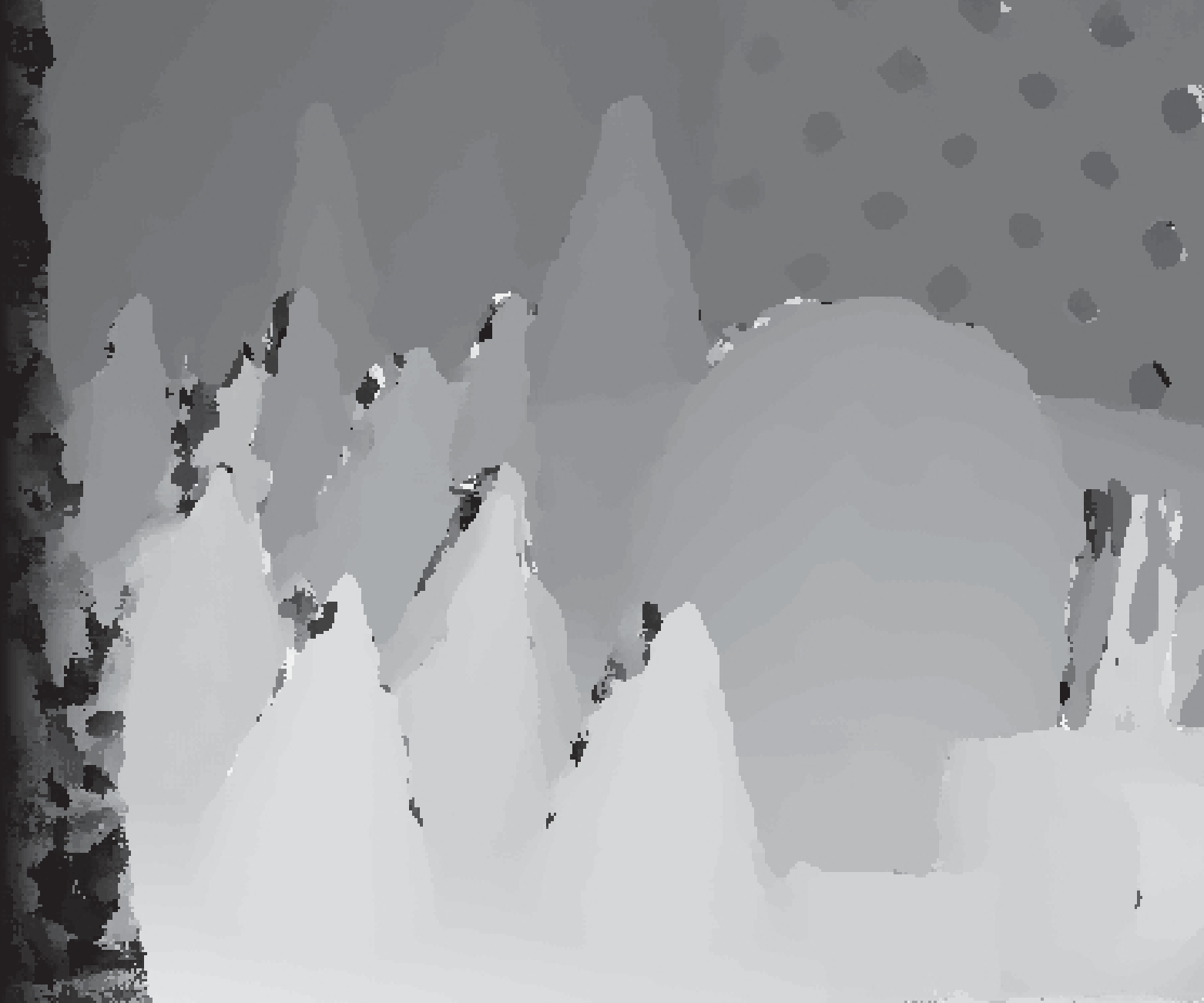}}
  \subfigure[]{\label{fig:cones_ld}\includegraphics[width=0.1\textwidth]{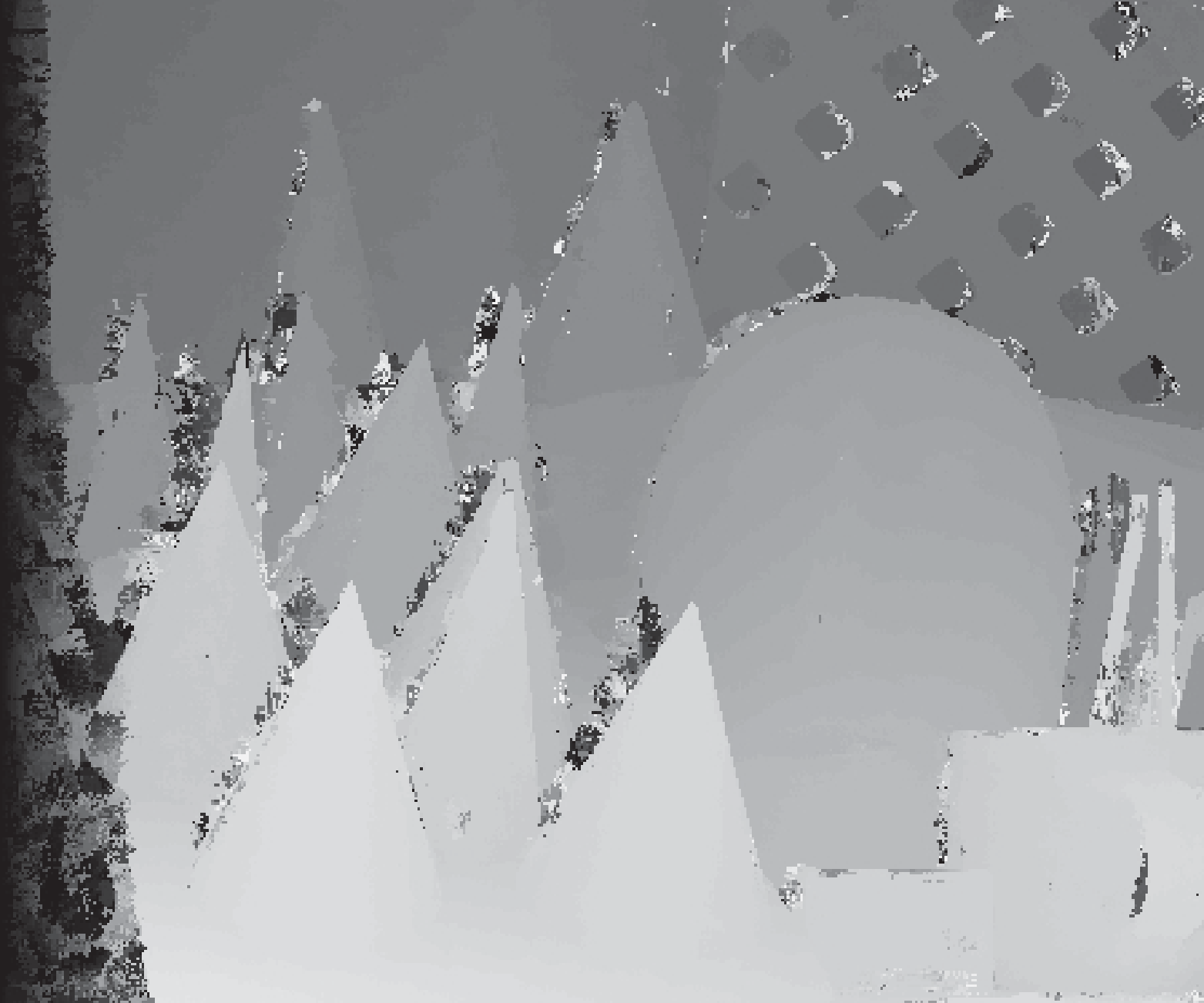}}
  \subfigure[]{\label{fig:cones_lp}\includegraphics[width=0.1\textwidth]{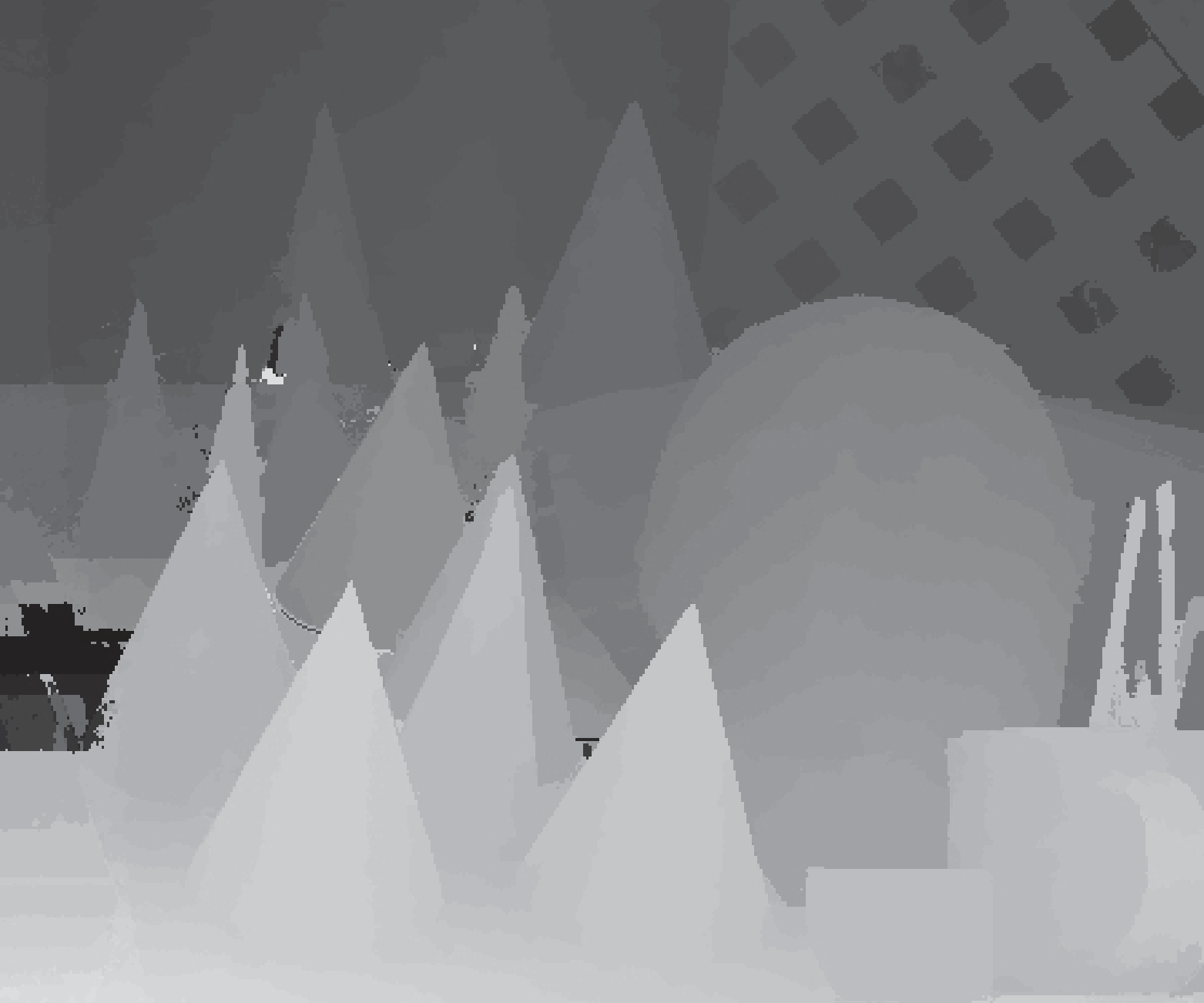}}
   \caption{Depth results of BSM after different stages. (a) is the left view of input image pair (Cones dataset \cite{middlebury}); (b)--(d) are depth results after cost computation, cost aggregation and depth refinement}
  \label{fig:cones}
  \vspace{-23pt}
\end{figure}

Stereo matching, which is to estimate depth or disparity map from two rectified images (left/right view), is a traditional problem in computer vision. It has wide applications in many areas including image-based rendering, robot navigation, etc. State-of-the-art stereo matching algorithms can generate reasonably good depth maps for images under ideally-configured illumination \cite{middlebury}. However, real stereo images usually have radiometric differences between left and right views, making stereo matching much more difficult \cite{hirschmuller_07}. Scharstein et al. gave a detail taxonomy and evaluation of stereo matching algorithms in \cite{scharstein_02} and according to \cite{ scharstein_02}, most stereo methods mainly consist of four steps: \emph{cost computation}, \emph{cost aggregation}, \emph{depth optimization} and \emph{depth refinement} (Figure \ref{fig:cones}). Based on different strategies adopted in depth optimization, stereo methods can be mainly classified into two categories: local methods and global methods.

In stereo matching, cost computation and aggregation is to construct a matching cost volume $C(x,d)$, where $x$ represents an image pixel from one view, $d$ represents one of all possible disparity values for $x$ and $C(x,d)$ is the matching cost when assigning disparity $d$ to pixel $x$. Cost volume largely determines the performance of stereo algorithms in  both local methods and global methods. In local methods, final disparity assignment for pixel $x$ is calculated using winner-take-all (WTA) scheme:
\vspace{-8pt}
\begin{equation}
\label{eqn:wta}
    d_{x} = \mathop{\arg\min}_{d \in D_d}{C(x,d)}
    \vspace{-6pt}
\end{equation}
where $d_{x}$ is the disparity for pixel $x$ and $D_d$ represents possible disparity ranges (in most of the cases, $D_d=[0,d_{max}-1]$). In global methods, the data term is constructed based on the cost volume. Thus, up-to-date surveys of stereo matching \cite{gong_07,hirschmuller_07,tombari_08} also focus on different approaches applied in cost computation and aggregation steps.

In this paper, we propose a novel cost computation and aggregation approach for stereo matching. By combining BRIEF feature descriptor \cite{calonder_10} with a novel binary mask, our method's cost volume is constructed using bitwise operations on binary strings. Then we adopt this approach into WTA scheme, resulting in a local stereo method called \emph{binary stereo matching} (BSM). We have conducted two experiments to demonstrate the performance of our algorithm. We first test our algorithm on Middlebury benchmark \cite{middlebury} and BSM has comparable performance compared with state-of-the-art methods \cite{de-maeztu_11,hosni_09,bleyer_11,rhemann_11} with slightly less time consumption. Furthermore, we also test BSM on datasets with radiometric differences \cite{hirschmuller_07}. Experimental results show that BSM is robust to radiometric difference especially to exposure changes, which demonstrates that BSM is much more suitable for unconstrained environment where illumination may have large variations. Another point we want to mention is that the core algorithm of BSM is based on binary and integer computations, so it will still be fast on embedded or mobile devices which do not have powerful floating point units.

The rest of this paper is organized as follows. We review some state-of-the-art local methods in Section \ref{sec:state}. Then our cost computation and aggregation approach together with BSM are explained in Section \ref{sec:pa}. Experimental results and analysis are given in Section \ref{sec:exp}. Finally we draw conclusion in Section \ref{sec:con}

\vspace{-10pt}
\Section{State-of-the-Art}
\label{sec:state}
\vspace{-10PT}

In this section, we briefly review state-of-the-art local methods. Bleyer et al. estimate a 3D plane at each pixel by applying PatchMatch \cite{barnes_09} into stereo matching and their method is currently top-performer among local methods. Hosni et al. \cite{hosni_09} aggregate matching cost by computing geodesic distance from all pixels to the window's center. De-Maeztu et al. \cite{de-maeztu_11} and Rhemann et al. \cite{rhemann_11} both adopt guided filter \cite{hekaiming_10} for cost aggregation and have speed advantages comparing to traditional local methods like \cite{yoon_06}. Detailed review of other traditional local methods is proposed in \cite{gong_07,hirschmuller_07,tombari_08}. Overall, most of state-of-the-art local methods use absolute pixel intensity difference for composing cost volume so that their performances drop dramatically under radiometric differences. Some methods explicitly handle radiometric differences like rank and census transform \cite{zabih_94}, however, their performances on normal images are not so good compared with state-of-the-art local methods \cite{ hirschmuller_07}. Our binary stereo matching algorithm not only achieves comparable performance with state-of-the-art methods but is also robust to the radiometric differences (especially to exposure changes).

\vspace{-10pt}
\Section{Proposed Approach}
\label{sec:pa}
\vspace{-10pt}

In this section we will explain our approach in detail. Our binary stereo matching algorithm also follows the classical four steps as stated before.

In cost computation, our approach is completely different with traditional local methods. We directly introduce BRIEF descriptor \cite{calonder_10} into cost computation. Thus, BRIEF descriptor $B(x)$ is calculated for every pixel $x$ in the input image pair. According to \cite{calonder_10}, $B(x)$ is defined as:
\begin{equation}
  \label{eqn:brief}
  B(x) = \sum_{1 \leq i \leq n}2^{i-1}\tau(p_i,q_i)
\end{equation}
where $\langle p_1, q_1 \rangle, \langle p_2, q_2 \rangle, \ldots, \langle p_n, q_n \rangle$ are $n$ pairs of pixels. Each pair $\langle p_i, q_i \rangle$ is sampled by an isotropic Gaussian distribution in a $S \times S$ window, which is centered on pixel $x$. And $\tau(p_i,q_i)$ is a binary function which is defined as:
\begin{equation}
    \label{eqn:binary}
    \tau(p_i,q_i)=\left\{\begin{array}{r@{\quad:\quad}l}
        1 & I(p_i) > I(q_i) \\0 & I(p_i) \leq I(q_i)\end{array}\right.
\end{equation}
where  $I(x)$ denotes the intensity of pixel $x$. After calculating the descriptor, i.e. a binary string for each pixel, the cost volume is constructed as:
\begin{equation}
    \label{eqn:cc}
    C(x,d)=\Vert\;B(x)\;\mathbf{XOR}\;B(x_d)\;\Vert_{1}
\end{equation}
where $x_d$ is the corresponding pixel of $x$ with disparity $d$ in another view, $\mathbf{XOR}$ is a bitwise xor-operation. In short, $C(x,d)$ measures the hamming distance between two binary strings.

Directly using BRIEF for stereo matching is a straightforward thought, which can be implemented by adopting $C(x,d)$ in (\ref{eqn:cc})  into WTA strategy as mentioned in (\ref{eqn:wta}). However, this naive approach will lead to the well-known edge-fattening problem as shown in Figure \ref{fig:cones_noca}. To solve edge-fattening problem, we invent a novel cost aggregation method by introducing another binary string which we call binary mask. Firstly, we define a weight function for pixel pair $\langle p_i, q_i \rangle$ in (\ref{eqn:brief}) as:
\begin{equation}
  \label{eqn:weight}
  w(x,p_i,q_i)=\max( SAD(x,p_i),SAD(x,q_i) )
\end{equation}
where $SAD(x,y) = \sum_{c \in [L,A,B]}|I_c(x)-I_c(y)|$ is the sum of absolute difference between two pixels in the CIELAB color space. Then we get our bitwise mask function for a given pair $\langle p_i, q_i \rangle$ as:
\begin{equation}
  \delta(x,p_i,q_i)=\left\{\begin{array}{r@{\quad:\quad}l}
  1 & w(x,p_i,q_i) \leq T \\0 & w(x,p_i,q_i) > T\end{array}\right.
\end{equation}
where $T$ is set to be the quarter smallest value in the sequence $w(x,p_1,q_1), w(x,p_2,q_2), \ldots, w(x,p_n,q_n)$. Finally, we can use this mask function to compose a binary mask:
\begin{equation}
  \label{eqn:mask}
  \Phi(x)=\sum_{1 \leq i \leq n}2^{i-1}\delta(x,p_i,q_i)
\end{equation}
Consequently $\Phi(x)$ is the proposed binary mask for cost aggregation. Incorporating the binary mask into (\ref{eqn:cc}), the new cost volume is defined as:
\begin{equation}
    \label{eqn:ca} C(x,d)=\Vert B(x)\;\mathbf{XOR}\;B(x_d)\;\mathbf{AND}\;\Phi(x)\Vert_1
\end{equation}

According to the definition of the binary mask, it will preserve those pixel pairs who have similar depth with window's center. After adopting our cost aggregation method, the edge-fattening effect is ideally removed as shown in Figure \ref{fig:cones_ld}.

Since local WTA strategy cannot handle occluded area, there are a large amount of errors in this region (as shown in Figure \ref{fig:cones_ld}). Besides, a small amount of random errors appear at non-occluded region due to mismatch. Thus, like other local methods, a depth refinement step is needed for removing these errors \cite{scharstein_02}. In our BSM algorithm, we propose a voting-based depth refinement method. Firstly, a left/right check using two-view depth maps is performed \cite{bleyer_11} to classify depth results into two categories: \emph{valid} and \emph{invalid}. Then we just refine those invalid pixels' depth using a voting schema. For an invalid pixel $x$, its refined depth is calculated as:
\begin{equation}
  \label{eqn:dp}
  d_x=\mathop{\arg\max}_{d \in D_d}{W(x,d)}
\end{equation}
where $W(x,d)$ represents accumulated weight voted from valid pixels, which is defined as:
\begin{equation}
  \label{eqn:vote} W(x,d)=\sum_{p}{\exp(-(\frac{c(x,p)}{\lambda_{c}}+\frac{e(x,p)}{\lambda_{e}}))}
\end{equation}
where $p$ represents a valid pixel with disparity $d$. And the accumulated weight is calculated according to bilateral filter
 \cite{tomasi_98} where $c(x,p)$ and $e(x,p)$ are distances between two pixels in color and Euclidean space respectively. In our implementation, parameters for bilateral filter are set as: $\lambda_{c}=9, \lambda_{e}=16$. As shown in Figure \ref{fig:cones_lp}, errors are corrected after refinement.

\vspace{-10pt}
\Section{Experimental Results and Analysis}
\label{sec:exp}
\vspace{-10pt}

In this section, we conduct a set of experiments to demonstrate the effectiveness of the proposed stereo algorithm.

\vspace{-8pt}
\SubSection{Comparison with state-of-the-art}
\label{subsec:com}
\vspace{-8pt}

\begin{table}[t]
  \caption{Depth results evaluation.}
  \begin{center}
  \label{tab:com}
  \begin{tabular}{c|c}
        \hline
        Methods & Average Error(\%)
        \\
        \hline
        PatchMatch\cite{bleyer_11} & 4.59
        \\
        \textbf{BSM} & \textbf{5.42}
        \\
        CostFilter\cite{rhemann_11} & 5.55
        \\
        P-LinearS\cite{de-maeztu_11} & 5.68
        \\
        GeoSup\cite{hosni_09} & 5.80
        \\
        \hline
      \end{tabular}
  \end{center}
  \vspace{-24pt}
\end{table}

\newcommand{\rb}[1]{\raisebox{1.3ex}[0pt]{#1}}
\begin{table}[b]
  \vspace{-24pt}
  \caption{Speed evaluation.}
  \begin{center}
  \label{tab:speed}
  \begin{tabular}{c|c|c}
        \hline
                     & Processor     & Running \\
        \rb{Methods} & Frequency(Hz) & Time(s)
        \\
        \hline
        \textbf{BSM} & \textbf{2.67} & \textbf{50}
        \\
        P-LinearS\cite{de-maeztu_11} &  2.13 & 94
        \\
        \hline
      \end{tabular}
  \end{center}
\end{table}

To compare with state-of-the-art local methods, we test BSM on standard datasets from Middlebury website \cite{middlebury}. In our implementation, we set $n=4096$, $S=26$, and the standard variance for the isotropic Gaussian distribution to be $4$. We use the same parameters for all datasets. Comparison between our algorithm with other methods is presented in Table \ref{tab:com}. We just list average error rate here to save space and detailed comparison can be found from our submission on Middlebury website \cite{middlebury}. In addition, we give a rough comparison of BSM's computational time with the fastest local methods. Since different methods are implemented under different platforms and there are many programming techniques (parallel computing or not) which affects the speed of the algorithm, we only choose P-LinearS \cite{de-maeztu_11} as representative of the fastest local methods for comparison, which has similar hardware configuration and implementation technique with our method (both algorithms are tested with one core and one process). Table \ref{tab:speed} shows test result on Teddy dataset, which involves the largest computational cost among all four datasets. Our algorithm is  much faster than \cite{de-maeztu_11} using a slightly better CPU.

\vspace{-8pt}
\SubSection{Resistance to radiometric differences}
\label{subsec:radio}
\vspace{-8pt}


\begin{figure}
  \centering
  \subfigure[Different exposure]
  {\label{fig:exposure}\includegraphics[width=0.2\textwidth]{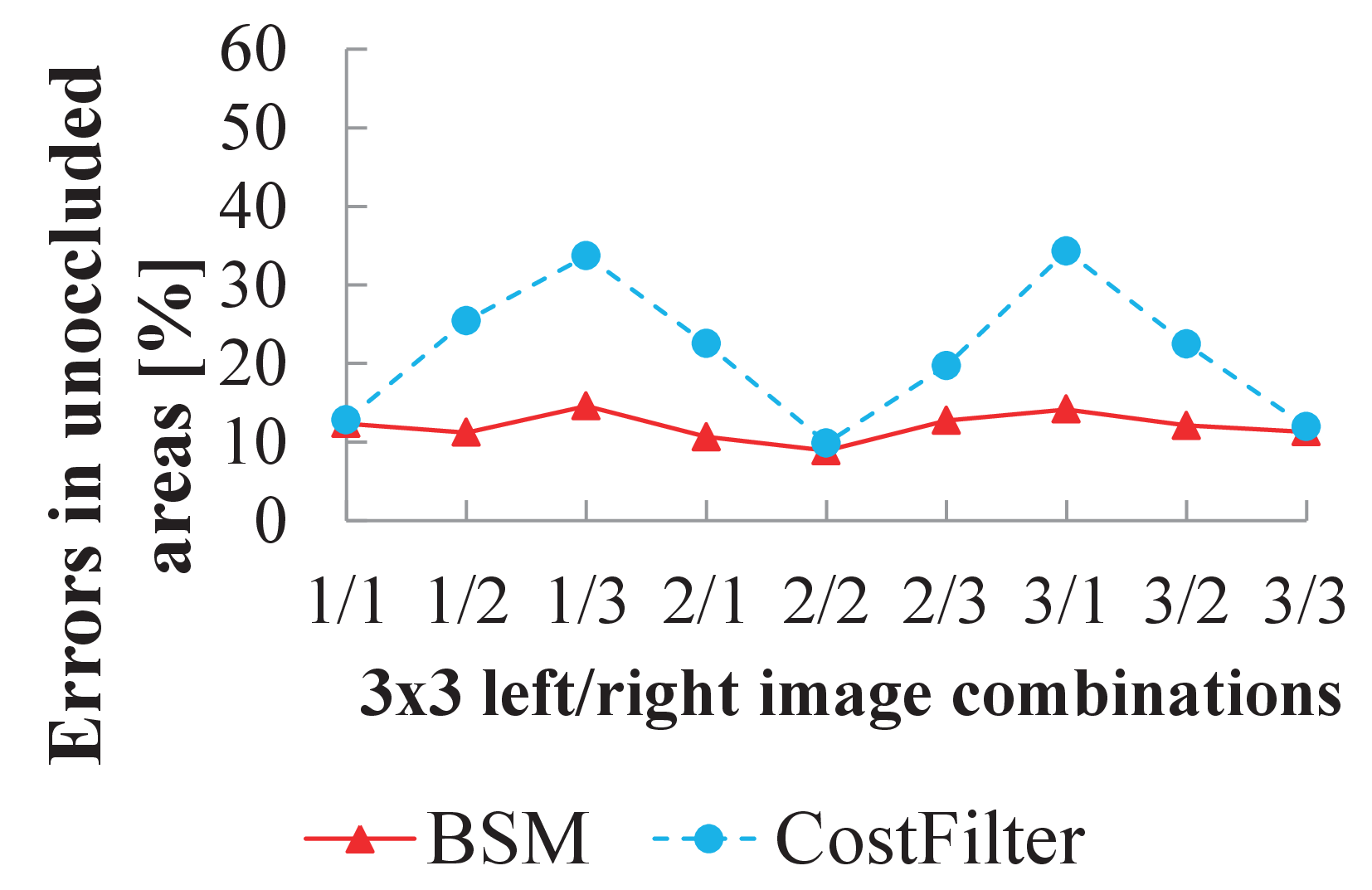}}
  \subfigure[Different lighting]
  {\label{fig:lighting}\includegraphics[width=0.2\textwidth]{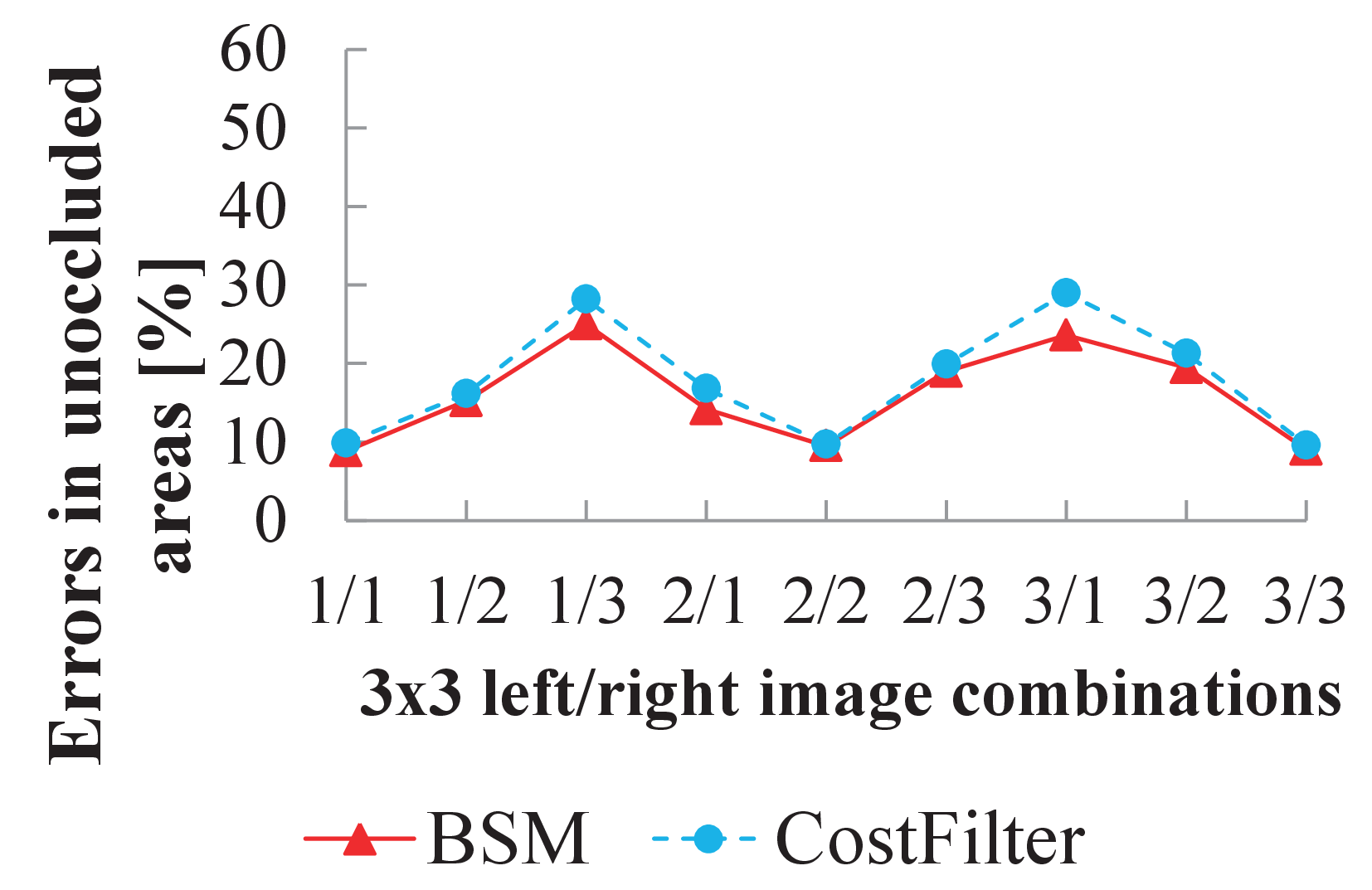}}
  \caption{Matching 3$\times$3 left/right image combinations that differ in exposure or lighting conditions.}
  \label{fig:radio}
  \vspace{-20pt}
\end{figure}

The traditional four datasets from Middlebury are configured under ideal illumination. For real images, there may be some radiometric changes. Thus, Hirschmuller et al. proposed new datasets incorporating exposure and lighting changes and evaluated some stereo methods on these datasets \cite{hirschmuller_07}. These new datasets give rectified image pairs under three different exposures and three lighting conditions. We conduct the same experiment as that mentioned in \cite{hirschmuller_07} and use the same evaluation methodology. For comparison, we also test CostFilter\cite{ rhemann_11} on these datasets (using the source code provided by the author). As shown in Figure \ref{fig:radio}, comparing to CostFilter\cite{rhemann_11}, BSM has much better performance under exposure changes. As for lighting changes, both BSM and CostFilter do not show good performance. As stated in \cite{hirschmuller_07}, it is of great difficult to handle local radiometric changes caused by changing the location of the light sources.

\vspace{-5pt}
\SubSection{Influence of descriptor length}
\label{subsec:inf}

\vspace{-15pt}
\begin{figure}[h]
    \centering
    \subfigure[Quality]{
      \label{fig:quality}\includegraphics[width=0.2\textwidth]{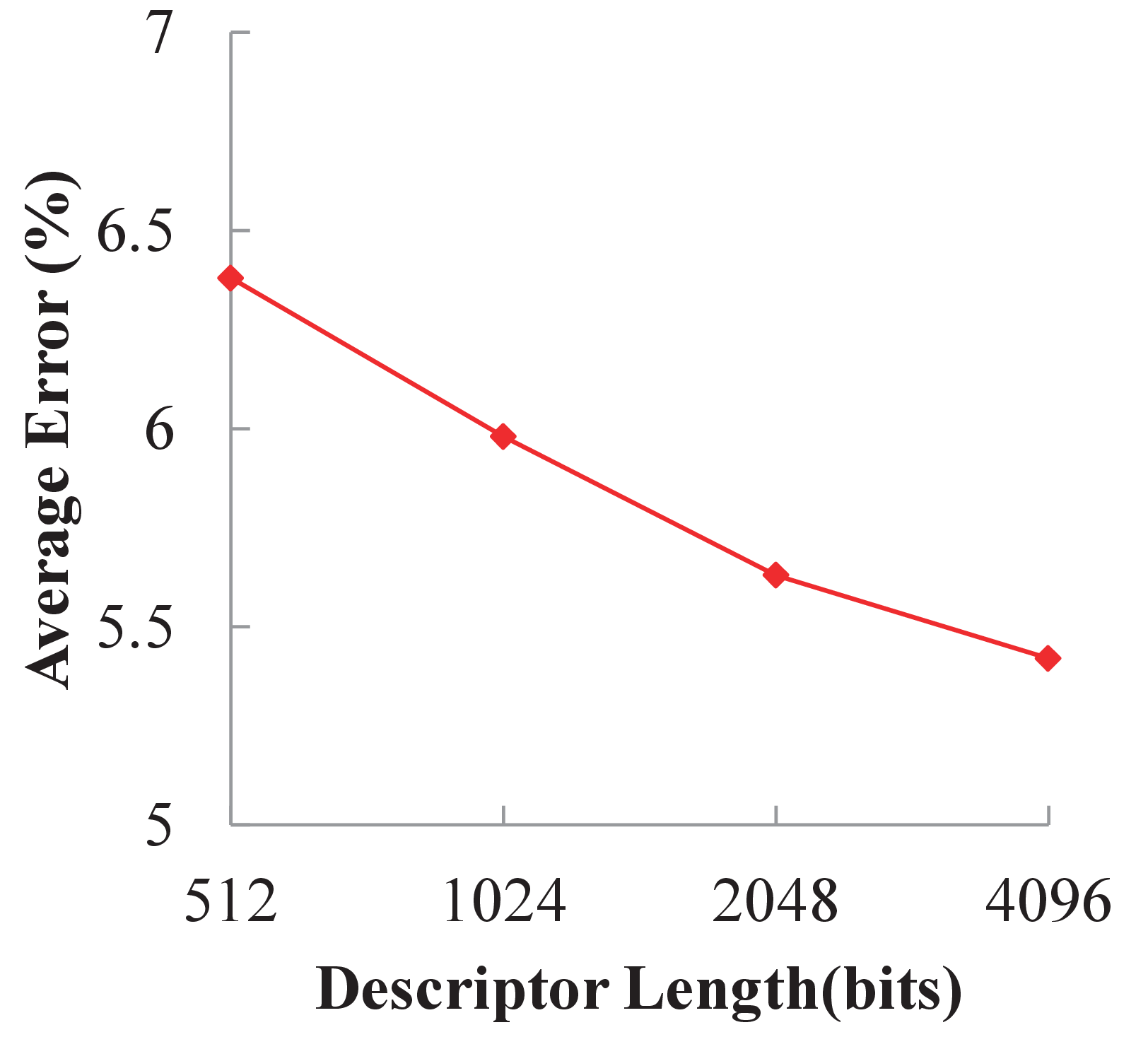}
    }
    \subfigure[Speed]{
      \label{fig:speed}\includegraphics[width=0.2\textwidth]{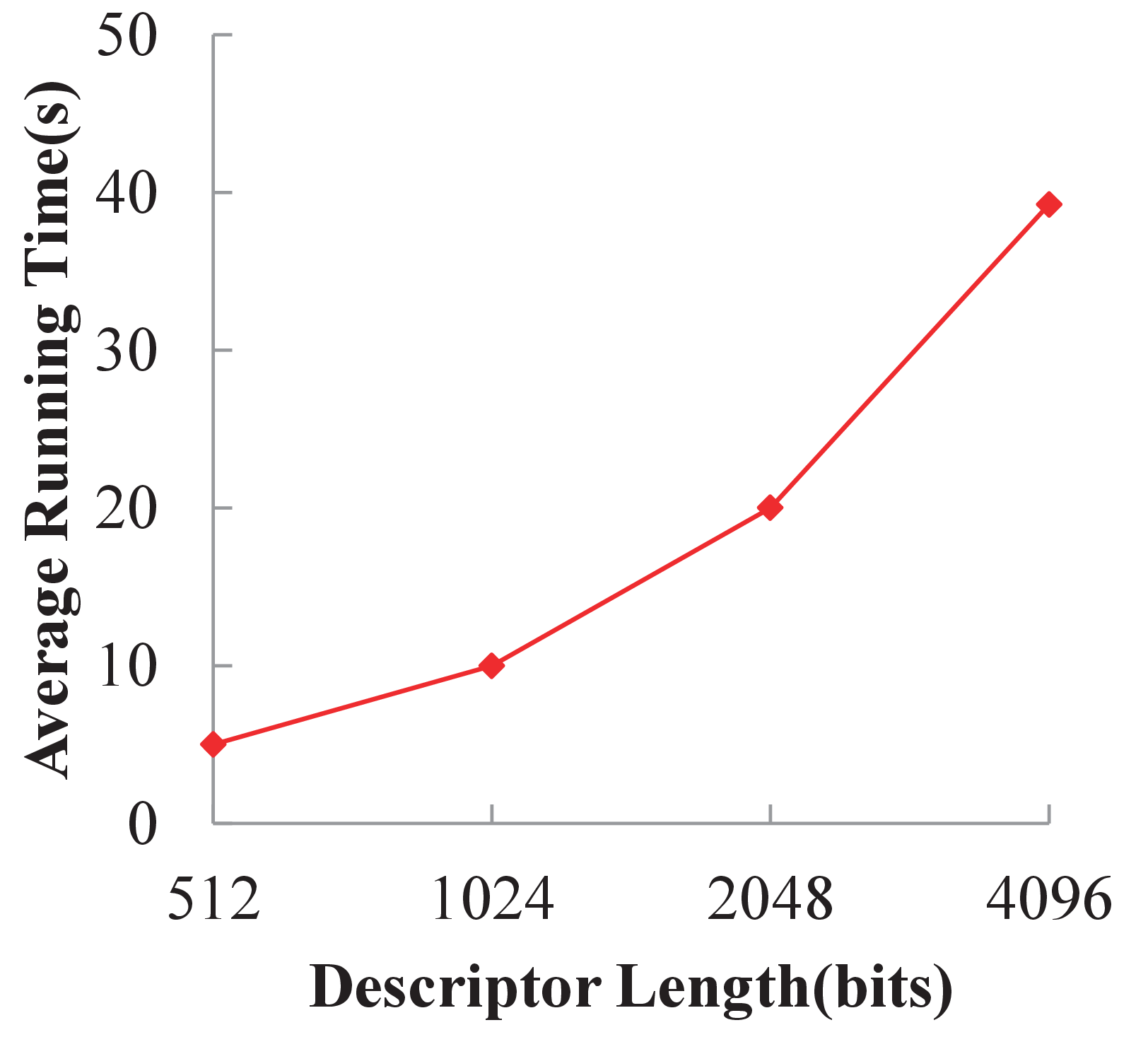}
    }
    \caption{The influence of descriptor length on the performance of BSM.}
    \label{fig:inf}
    \vspace{-18pt}
\end{figure}

It is easy to be proved that our algorithm's computational complexity is $O(whnd_{max})$ ($w$ and $h$ are image width and height respectively). Thus running speed of BSM is mainly determined by the descriptor length $n$. Also, the descriptor length affects depth map's quality because longer descriptor implies a dense sampling. To show the influence of descriptor length on the performance of BSM, we test BSM with different $n$ on the traditional four datasets from Middlebury. Experimental result is shown in Figure \ref{fig:inf}, which is consistent with the analysis above. This interesting property of BSM makes it easy to gain different tradeoff between speed and quality in different scenarios.

\vspace{-10pt}
\Section{Conclusion}
\label{sec:con}
\vspace{-10pt}

In this paper, a novel cost computation and aggregation approach for stereo matching is proposed. Combining our cost computation and aggregation approach with WTA strategy, we design a new local stereo method called binary stereo matching. The proposed algorithm is mainly based on binary and integer computations, so it is fast and fits for embedded or mobile devices. Experimental results show that BSM has a better performance either on traditional stereo datasets or on new datasets with radiometric differences. In the future, we will definitely incorporate our cost computation and aggregation approach into global optimization and implement BSM on GPU to achieve real-time matching.

\vspace{-10pt}
\Section{Acknowledgement}
\label{sec:ack}
\vspace{-10pt}

This work has been partially supported by the Development Plan of
China (973) under Grant No. 2011CB302206, the National Natural Science
Foundation of China under Grant No. 60833009/60933013, and the Research Grant of Tsinghua-Tencent Joint Lab.




\scriptsize \tiny
\bibliographystyle{latex12}
\bibliography{allbib}

\end{document}